\documentclass{article}
\usepackage{spconf}
\usepackage{amssymb}
\usepackage{graphicx}
\usepackage{graphicx,color}
\usepackage{amssymb, array}
\usepackage[margin=0.750in]{geometry}

\usepackage[cmex10]{amsmath}

\usepackage{array}

\usepackage{mdwmath}
\usepackage{mdwtab}

\usepackage{tikz}
\usepackage{epigraph}

\usepackage{eqparbox}

\usepackage{cite}
\usepackage[cmex10]{amsmath}
\usepackage{amssymb}
\usepackage{amsthm} 
\usepackage{amsmath}
\usepackage{graphicx}
\usepackage{subfigure}
\usepackage{color,soul} 
\usepackage{mathtools} 
\usepackage{forloop}  
\usepackage{algorithm,algpseudocode}
\usepackage{multicamfigs}
\usepackage{braket} 
\usepackage{tabularx}
\usepackage{tkz-euclide}
\usetkzobj{all}
\usepackage{pgfplots}
\ifpdf\else
\usepackage{pstricks,pst-node,pst-plot,pst-circ}
\psset{pgffunctions}
\fi
\usepackage[font=footnotesize]{subfig}

\usepackage{url}

\DeclareMathOperator{\sign}{sign}
\DeclarePairedDelimiter\floor{\lfloor}{\rfloor} 

\usepackage{multicamfigs}

\def\x{{\mathbf x}}

\graphicspath{{images/}}
\title{SHAPE: Linear-Time Camera Pose Estimation With Quadratic Error-Decay}
\name{
	Alireza Ghasemi \hspace*{20pt} Adam Scholefield\hspace*{20pt} Martin Vetterli 
	\thanks{This work was supported by the ERC Advanced Grant---Support for Frontier Research---SPARSAM Nr: 247006.}
	\thanks{A. Ghasemi is additionally supported by a Qualcomm Innovation Fellowship.}
}
\address{\vspace*{-0pt}
School of Computer and Communication Sciences\\
\'{E}cole Polytechnique F\'{e}d\'{e}rale de Lausanne\\
	}
\small
\begin{document}
%
\maketitle

\begin{abstract}
We propose a novel camera pose estimation or perspective-n-point (PnP) algorithm, based on the idea of consistency regions and half-space intersections. Our algorithm has linear time-complexity and a squared reconstruction error that decreases at least quadratically, as the number of feature point correspondences increase. 

Inspired by ideas from triangulation and frame quantisation theory, we define consistent reconstruction and then present SHAPE, our proposed consistent pose estimation algorithm. We compare this algorithm with state-of-the-art pose estimation techniques in terms of accuracy and error decay rate. The experimental results verify our hypothesis on the optimal worst-case quadratic decay and demonstrate its promising performance compared to other approaches.

\end{abstract}


%
%

%
%
\section{Introduction}
Camera pose estimation, or the perspective-n-point (PnP) problem, aims to determine the pose (location and orientation) of a camera, given a set of correspondences between 3-D points in space and their projections on the camera sensor~\cite{zheng2013revisiting}. The problem has applications in robotics, odometry~\cite{quan1999linear}, and photogrammetry, where it is known as space resection~\cite{haralick1994review}.

In the simplest case, one can use an algebraic closed-form solution to derive the camera pose from a set of minimal 3D-to-2D correspondences. Usually, three correspondences are used and hence these algorithms are called perspective-3-point or P3P methods~\cite{kneip2011novel,ferraz2014very}. 

When there is a redundant set of points available (more than three), the most straightforward solution is to use robust algorithms, such as RANSAC, which run P3P (or its variants) on minimal subsets of correspondences~\cite{fischler1981random}. However, such algorithms suffer from low accuracy, instability and poor noise-robustness, due to the limited number of points. 

An alternative approach is to directly estimate the camera pose, using an objective function, such as the $\ell_2$-norm of the reprojection error, defined over all available point correspondences~\cite{lu2007fast,kahl2008practical}. 

Minimisation of the $\ell_2$-norm leads to the maximum likelihood estimator, if we assume a Gaussian noise model. However, the main drawback of the $\ell^2$-norm is that its resulting cost function is non-convex and usually has a lot of local minima~\cite{hartley2007optimal}. This forces us to use iterative algorithms that are reliant on a good initialisation~\cite{DBLP:journals/pr/KangWY14}. 

The shortcomings of the $\ell_2$-norm have encouraged researchers to consider using other norms, such as the  
$\ell_\infty$-norm~\cite{kahl2008multiple}. The main advantage of the $\ell_\infty$-norm is that its minimisation can be formulated as a quasi-convex problem and solved using Second-Order Cone Programming (SOCP)~\cite{hartley2007optimal,mittelmann2003independent}. This leads to a unique solution, however SOCP techniques are computationally demanding and rely on the correct tuning of extra parameters~\cite{dai2009smooth}. 

There is an interesting, well known, duality between pose estimation and triangulation, which allows common algorithms to be used for both problems \cite{aastrom2007approach,kahl2008practical}. Triangulation estimates the location of a point given its projection in a number of calibrated cameras~\cite{hartley1997triangulation}. Various triangulation algorithms exist, which once again mostly relying on minimising the reprojection error~\cite{hartley2007optimal}. To see the duality, notice that in both cases we have a set of projections and we want to estimate the location of an object of interest; i.e., the camera, in pose estimation, and the point, in triangulation. 


In this paper, we propose a fundamentally novel approach to the camera pose estimation problem. Under certain assumptions, this leads to the optimal estimate for both the camera location and orientation, and a consistency region, where the true camera pose must lie. Our algorithm has a linear time-complexity, allowing it to be used efficiently with a large number of points. Moreover, exploiting the duality between camera pose estimations and triangulation, we use our earlier work \cite{ghasemi2015accuracy,ghasemi2015pami} to show that the expected error decays at least quadratically as we increase the number of available point correspondences.

In the rest of this paper, we formally define the problem and our imaging setup. For simplicity, we limit our discussion to 1-D vision systems, in which points in $\mathbb{R}^2$ are mapped to points on a 1-D image sensor. Although this special case is independently important, e.g. in developing autonomous guided vehicles or planar motion~\cite{aastrom2007approach}, the extension to 3-D vision is straightforward. In addition, we assume that the points' locations are known exactly.

After defining the problem setup, we propose our algorithm, coined SHAPE (Sequential Half-Space Aggregation for Pose Estimation). We start with the simpler case of known camera orientation (i.e. estimating only the location of the camera) and then extend the algorithm to compute the orientation as well. Finally, we present experimental results comparing our algorithm to state-of-the-art techniques. The results verify our hypothesis that the worst-case error decay rate of our algorithm is quadratic and demonstrate its promising performance compared to other approaches.  
\section{Problem Setup}

We now introduce the digital pinhole camera model which will be assumed throughout this paper. 
Our aim is to estimate the orientation $\theta$ and location $\textbf{t}=(t_x,t_z)$ of a camera having a resolution of $N$ pixels, given $M$ 2D-to-1D correspondences between points $\textbf{s}_i\in\mathbb{R}^2$, and their pixelised projections $q_i$ on the camera. 

As depicted in Fig. \ref{fig:acquisition_setup}, we assume that the $i$-th point source $\textbf{s}_i$, is projected to the position $p_i$ on the camera's image plane before being quantised to $q_i$, the centre of the corresponding pixel. As well as modelling the finite pixel width, $q_m$ also models the finite sensor width, by obtaining the value $\mathtt{\sim}$ if the projected point lies outside the field of view. Later we will consider algorithms that take the quantised feature projections $q_i$ and estimate the pose $(t_x,t_z,\theta)$ of the camera.

As shown in Fig. \ref{fig:acquisition_setup}, the camera is centred at $\boldsymbol{t}$ and orientated $\theta$ radians anti-clockwise from the global coordinate system. With this notation, the projected point $p_i$ is given by\footnote{Here, and in the rest of this paper, we have assumed perspective projection, which is standard in most imaging applications. However, we can derive similar results for other camera models. In fact, orthogonal projection have been extensively studied in the quantisation literature \cite{goyal1998quantized,cvetkovic1999source} and also in image processing~\cite{ghasemi2015accuracy}.}
\begin{align}
p_i = f\frac{(s_{i,x} - t_x)\cos \theta+(s_{i,z} - t_z)\sin \theta}{(s_{i,z} - t_z)\cos \theta-(s_{i,x} - t_x)\sin \theta}.\label{eq:projection}
\end{align}
Here, $f$ is the focal length of the camera and $(s_{i,x},s_{i,z})$ and $(t_x,t_z)$ are the coordinates of the $i$-th point and camera centre, respectively, with respect to a global coordinate system. 

The quantised point, $q_i$, is given by $q_i = Q_\Lambda(p_i)$,
where $Q_\Lambda$ is the quantisation function defined as\footnote{This definition is valid when $N$ is even. For odd $N$, \hbox{$Q_\Lambda(y)=\begin{cases}
		\sign(y)\floor*{ \frac{|y|}{w} +\frac{1}{2}} w &\quad-\frac{\tau}{2} \leq y \leq \frac{\tau}{2},\\
		\,\mathtt{\sim} &\quad\text{otherwise}.
		\end{cases}$}}
\begin{equation}
Q_\Lambda(y)=\begin{cases}
\floor*{\frac{y}{w}} w + \frac{w}{2} &\quad-\frac{\tau}{2} \leq y \leq \frac{\tau}{2},\\
\,\mathtt{\sim} &\quad\text{otherwise}.
\end{cases}
\label{eq:quant}
\end{equation}
In \eqref{eq:quant}, $\Lambda=\left\{\tau,w\right\}$ encapsulates the sensor width $\tau$ and the pixel width $w=\frac{\tau}{N}$, which define the quantisation error. 


In the rest of the paper, we will be interested in PnP algorithms that take the pixelised projections $q_i$ and the true locations of the $M$ point sources $\boldsymbol{s}_i$, and produce an estimate of the camera pose ($\boldsymbol{t}$, $\theta$).


\tikzset{
  pics/carc/.style args={#1:#2:#3}{
    code={
      \draw[pic actions] (#1:#3) arc(#1:#2:#3);
    }
  }
}

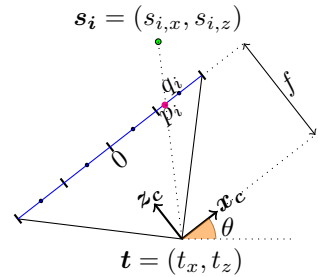
\begin{figure}[t!]
\captionsetup[subfloat]{position=top,labelformat=empty}
	\centering
	\vfil
			\begin{tikzpicture}[scale = 1.5]
			\renewcommand{\rad}{2.5}
			\renewcommand{\ns}{4}
			\renewcommand{\nc}{7}
			\renewcommand{\fov}{90.00}
			\newcommand{\foclen}{1.03}
			
			\begin{scope}[shift={(4.8,2),scale=2}]
			
			\coordinate (C1) at (.5,.25);
			\begin{scope}[shift={(C1)},rotate=38]
			\norprojvertonlycamnopix
			\pgfmathsetmacro\pr{0.6127}
			\pgfmathsetmacro\qu{0.7725}
			\coordinate (P1) at (\pr,\foclen);
			\node[rotate=38] at (\pr,0.93) {$p_i$};
			\node[rotate=38] at (\qu,1.13) { $q_i$};
			\draw[fill=magenta,color=magenta] (P1) circle (.025);
			\draw[ thick,->] (0,0) -- (.4,0);
			\draw[ thick,->] (0,0) -- (0,0.4);
			\node[rotate=38] at (0,0.47) {$\boldsymbol{z_c}$};
			
			\node[rotate=38] at (0,0.9) {$0$};
			
			\draw[<->] (1.5,0) -- (1.5,\foclen);
			\node[rotate=38] at (1.6,0.5) {$f$};
			
			\draw[-,dotted] (0,0) -- (1.5,0);
			\draw[-,dotted] (1.03,\foclen) -- (1.5,\foclen);
			
			\node[rotate=38] at (0.55,0) {$\boldsymbol{x_c}$};
			\end{scope}
			
			
			\coordinate  (A) at (0.5,0.25);
			\coordinate  (F) at (1.5,0.25);
			\coordinate  (E) at (1.29,0.86);
			\coordinate  (F2) at (1.5,0.1);
			
			\tkzMarkAngle[fill=orange,opacity=.5,size=.3](F,A,E)
			\tkzLabelAngle[pos=.42](E,A,F2){$\theta$}
			\draw[dotted] (A) -- (F);
			
			\coordinate (S) at (.29,2);
			\draw[fill=green](S) circle (.025);
			
			\draw[dotted] (S) -- (C1) ;
			
			\node[below] at (C1) {$\boldsymbol{t}=(t_x,t_z)$};
			\node[above] at (S) {$\boldsymbol{s_i}=(s_{i,x},s_{i,z})$}; 

			\end{scope}
			\end{tikzpicture}%
	\caption{The acquisition setup for a pinhole camera with a resolution of four pixels, acquiring a point source $\boldsymbol{s}$.}
	\label{fig:acquisition_setup}
\end{figure}


\section{SHAPE: Sequential Half-space Aggregation for Pose Estimation}

We now describe the proposed pose estimation algorithm, denoted SHAPE. We will first assume that we know the camera's orientation before considering the more general case. 

\subsection{Localising a Camera with Known Orientation}

We would like to see how each point source constrains the location of the camera. Given a quantised projection $q_i$, we know that the true projected point $p_i$ satisfies

\begin{equation}
q_i-\frac{w}{2} \leq p_i \leq q_i+\frac{w}{2},
\label{eq:pixel_bounds}
\end{equation}
where $w$ is the width of a pixel. Combining \eqref{eq:projection} with \eqref{eq:pixel_bounds} and rearranging yields

\begin{equation}
		a_it_x+b_it_z+c_i \geq 0,
\end{equation}
and
\begin{equation}
		a_i't_x+b_i't_z+c_i' \leq 0.
\end{equation}
Here $a_i$, $a_i'$, $b_i$, $b_i'$, $c_i$, and $c_i'$ are defined as
\small
\begin{align*}
	a_i&=f\cos\theta-(q_i+\frac{w}{2})\sin\theta, \\ 
	b_i&=f\sin\theta-(q_i+\frac{w}{2})\cos\theta, \\ 
	c_i&=(q_i+\frac{w}{2})(s_{i,z}\cos\theta+s_{i,x}\sin\theta)-fs_{i,x}\cos\theta-fs_{i,z}\sin\theta, \\
	a_i'&=f\cos\theta-(q_i-\frac{w}{2})\sin\theta, \\ 
	b_i'&=f\sin\theta-(q_i-\frac{w}{2})\cos\theta, \\ 
	c_i'&=(q_i-\frac{w}{2})(s_{i,z}\cos\theta+s_{i,x}\sin\theta)-fs_{i,x}\cos\theta-fs_{i,z}\sin\theta. \\
\end{align*}
\normalsize


Therefore, assuming we know the orientation of the camera, each point source constrains the camera location to lie between two half-spaces; i.e., within a semi-infinite triangle. 

When there are multiple points, then there is a semi-infinite triangular region for each point and the camera must lie in their intersection. Therefore, we need to compute the intersection of all half-spaces which produces a polygon where the camera centre $\boldsymbol{t}$ must lie. Every point within this polygon is consistent with the projections and has an equal chance of being the true camera location, as long as the feature detection accuracy is uniformly bounded within the pixel limits. SHAPE selects the point that leads to the minimum mean squared distance to all points inside the consistency polygon. This point is the centre of mass of the consistency region and can be computed very efficiently in constant time. 

Figure \ref{fig:hsi_example} visualises one example of applying our proposed algorithm to a camera positioning problem with three known points and their projections in a 6-pixel camera. 

\subsection{Simultaneous Estimation of Camera Location and Orientation}

When the camera orientation is unknown, we have an extra dimension which leads to a 3D solution space $(t_x,t_z,\theta)$. We previously analysed the $t_x$-$t_z$ slice for the true camera orientation and saw that there was a polygon, which led to estimates of the camera location that were consistent with the measurements. Let us now consider slices for an arbitrary angle $\theta$. As $\theta$ changes, each half space rotates around its point source. For many angles, there is no common intersection between the half spaces but there is a range of angles for which the intersection creates a polygon. Thus there is a 3D shape created by the union of all these slices, containing all estimates consistent with the measurements. 

We would like to find the centre of mass of this 3D shape, leading to an estimate of the location and orientation. This 3D region is neither a polytope nor convex and calculating its centroid is not trivial. We can find an accurate approximation by taking the weighted average of a finite number of slices. 

More precisely, suppose the camera orientations are discretised to $\Theta=\{0,\frac{2\pi}{k},\frac{4\pi}{k},\dots,2\pi\}$, and that, for every orientation $\alpha\in\Theta$, we have computed the location-consistency region $\mathcal{R}_{\alpha}$ and its centre of mass $\mathcal{C}(\mathcal{R}_{\alpha})$. We approximate the centre of mass of the 3-D consistency shape as
\begin{equation}
	(\hat{t}_x,\hat{t}_z,\hat{\theta})=\frac{\sum_{\alpha\in\Theta} \mathcal{A}(\mathcal{R}_{\alpha})\mathcal{C}(\mathcal{R}_{\alpha})}{\sum_{\alpha\in\Theta} \mathcal{A}(\mathcal{R}_{\alpha})},
	\label{eq:3dcentroid}
\end{equation}
where $\mathcal{A}(\mathcal{R}_{\alpha})$ is the area of the location-consistency region $\mathcal{R}_{\alpha}$. This is SHAPE's final estimate of the camera pose. 

\tikzset{
  pics/carc/.style args={#1:#2:#3}{
    code={
      \draw[pic actions] (#1:#3) arc(#1:#2:#3);
    }
  }
}

\begin{figure}[t!]
\captionsetup[subfloat]{position=top,labelformat=empty}
	\centering
	\vfil

		\begin{tikzpicture}[scale = .25]
		\renewcommand{\rad}{100}
		\renewcommand{\ns}{6}
		
		\renewcommand{\fov}{108.91} //f/s=10/28
		\newcommand{\foclen}{.01}
		\begin{scope}[rotate=180]
		\begin{scope}
		
		\clip(-23.5,-22) rectangle (5,5);		
		%
		\coordinate (C1) at (-20,0);
		\node[above] at (C1) {$\boldsymbol{s}_1$};
		\begin{scope}[shift={(C1)},rotate=-30]
		\pointconsistencyregionwithallpix{6};
		\end{scope}

		\coordinate (C2) at (-20,-20);
		\node[above] at (C2) {$\boldsymbol{s}_2$};
		\begin{scope}[shift={(C2)},rotate=-30]
		\pointconsistencyregionwithallpix{3};
		\end{scope}
		
		\coordinate (C3) at (0,-20);
		\node[above] at (C3) {$\boldsymbol{s}_3$};
		\begin{scope}[shift={(C3)},rotate=-30]
		\pointconsistencyregionwithallpix{1};
		\end{scope}
		
		%
		%

		

		\draw[blue,fill=lightgray, very thick,opacity=0.01]   (-7.768317,1.185896) --  (-9.559160,1.012269) --  (-10.427914,2.921902) --  (-5.982603,4.278844) --  (-7.768317,1.185896) ;
		\renewcommand{\foclen}{.75}
		\coordinate (C4) at (-8.345375,2.534551);
		\node[left] at (C4) {\vspace*{-50pt}$\hat{\boldsymbol{t}}$};
		\begin{scope}[shift={(C4)},rotate=150]
			\norprojvertcamandfocnopix{\foclen};
		\end{scope}
		
		

		
		
		\end{scope}
		
	
		\end{scope}
		\end{tikzpicture}%
	\caption{An example of using polygon intersections for estimating the camera pose. Boundaries of half-spaces are depicted as black lines. The intersection of half-spaces is the light-grey polygon. The centroid of this polygon is the reconstructed centre  $\hat{\boldsymbol{t}}$ of the camera. }
	\label{fig:hsi_example}
\end{figure}
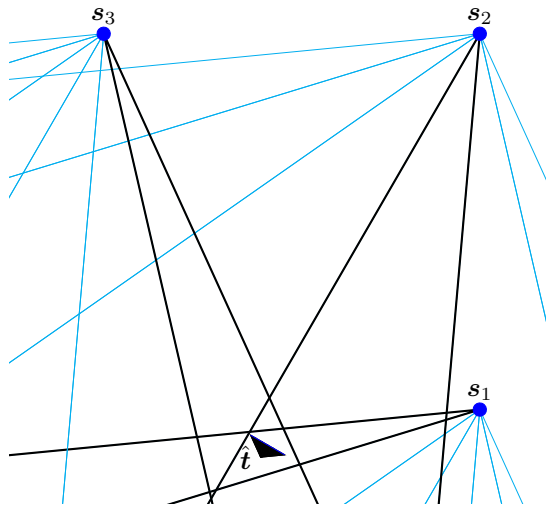

\subsection{Time Complexity of the SHAPE Algorithm}
The core part of the SHAPE algorithm is a series of $2M$ half-space intersection ($M$ is number of points). Since the area between half-spaces is a triangle in the finite case, we can solve this problem using polygon intersection algorithms. 

The worst-case time-complexity for polygon and half space intersection is $\mathcal{O}(M\log M)$, in the general case~\cite{preparata1979finding}. However, since the point-consistency regions form convex polygons, which can be intersected in $\mathcal{O}(V_1+V_2)$ time, where $V_1$ and $V_2$ are the number of vertices of the two polygons~\cite{toussaint1985simple}. Therefore, a series of intersections between convex polygons can be computed in $\mathcal{O}(MV_{max})$, where $V_{max}$ is the maximum number of vertices of an intermediate polygon. In our case, it can be intuitively shown that the number of vertices of intermediate polygons do not grow with the number intersected polygons, for any practical number of points. Therefore, we can bound $V_{max}$ and hereby reach a $\mathcal{O}(M)$, or linear time-complexity for our algorithm. 

\subsection{Error Decay Rate of the SHAPE Algorithm}
\label{sec:conv_rate}
We would like to know if our algorithm converges to the true latent value, as the number of point correspondences tend to infinity, and how fast the error decays. 

By adapting results from frame quantisation theory~\cite{goyal1998quantized}, we have recently shown a quadratic error decay rate for the dual triangulation problem with circular and linear camera arrays~\cite{ghasemi2015accuracy,ghasemi2015pami}. 



Triangulation with a linear camera array is equivalent to the PnP problem with collinear feature points and a known camera orientation. It follows that, if the shortest distance between the camera and the line where the points lie does not exceed $\frac{fb}{2w}$, where $b$ is the largest distance between the points, the error of the SHAPE algorithm decays quadratically as the number of feature point correspondences increases.

In general, however, if the points are distributed randomly, the error decays much faster, since the large consistency regions of the linear case are unlikely to occur. This is why we obtain a much faster error decay rate in practice, as can be seen in Fig. \ref{fig_or_loglog}. Moreover, the large consistency regions generated by collinear and coplanar point sets, explain the difficulties traditionally seen by PnP algorithms in these cases.

\section{Simulation Results}
\label{sec:experiments}
To assess our algorithm, we have randomly generated a set of camera poses and for each one randomly added points within the cameras field of view. Figure \ref{fig:expr_example} depicts a sample configuration used in the experiments. The camera has a resolution of $320$ pixels and a field of view of $90$ degrees.

\tikzset{
  pics/carc/.style args={#1:#2:#3}{
    code={
      \draw[pic actions] (#1:#3) arc(#1:#2:#3);
    }
  }
}

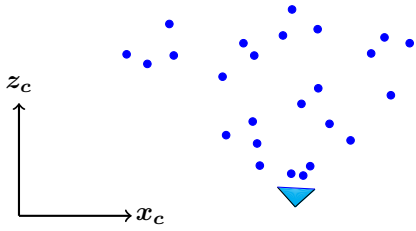
\begin{figure}[t!]
\captionsetup[subfloat]{position=top,labelformat=empty}
	\centering
	\vfil

		\begin{tikzpicture}[scale =.5]
		\renewcommand{\rad}{150}
		\renewcommand{\ns}{32}
		
		\renewcommand{\fov}{90} //f/s=10/28
		\newcommand{\foclen}{.5}
		\begin{scope}
		\begin{scope}
		
		%
		\coordinate (C1) at (-20,0);
		\begin{scope}[shift={(C1)},rotate=-30]
		\end{scope}

		\coordinate (C2) at (-20,-20);
		\begin{scope}[shift={(C2)},rotate=-30]
		\end{scope}
		
		\coordinate (C3) at (2.344553,0.238599);
		\begin{scope}[shift={(C3)},rotate=357]
			\norprojvertcamandfoc{\foclen};
		\end{scope}
		
		%
		%
		
		\newcommand{\cirrad}{.1}
		\draw[fill=blue,color=blue] (-2.137697,4.297372) circle (\cirrad);
		\draw[fill=blue,color=blue] (-1.586083,4.042499) circle (\cirrad);
		\draw[fill=blue,color=blue] (0.507970,2.146536) circle (\cirrad);
		\draw[fill=blue,color=blue] (1.403587,1.336470) circle (\cirrad);
		\draw[fill=blue,color=blue] (-0.884792,4.266876) circle (\cirrad);
		\draw[fill=blue,color=blue] (-1.001533,5.104591) circle (\cirrad);
		\draw[fill=blue,color=blue] (1.331484,1.925068) circle (\cirrad);
		\draw[fill=blue,color=blue] (0.413520,3.702200) circle (\cirrad);
		\draw[fill=blue,color=blue] (1.214441,2.508831) circle (\cirrad);
		\draw[fill=blue,color=blue] (0.968464,4.592852) circle (\cirrad);
		\draw[fill=blue,color=blue] (1.255844,4.264603) circle (\cirrad);
		\draw[fill=blue,color=blue] (2.238873,1.121273) circle (\cirrad);
		\draw[fill=blue,color=blue] (2.018762,4.799140) circle (\cirrad);
		\draw[fill=blue,color=blue] (2.260762,5.491002) circle (\cirrad);
		\draw[fill=blue,color=blue] (2.510891,2.979202) circle (\cirrad);
		\draw[fill=blue,color=blue] (2.941802,4.965943) circle (\cirrad);
		\draw[fill=blue,color=blue] (2.957738,3.393493) circle (\cirrad);
		\draw[fill=blue,color=blue] (2.557745,1.071596) circle (\cirrad);
		\draw[fill=blue,color=blue] (2.743264,1.320882) circle (\cirrad);
		\draw[fill=blue,color=blue] (3.257036,2.447586) circle (\cirrad);
		\draw[fill=blue,color=blue] (4.364665,4.320529) circle (\cirrad);
		\draw[fill=blue,color=blue] (4.719038,4.747368) circle (\cirrad);
		\draw[fill=blue,color=blue] (5.388435,4.593283) circle (\cirrad);
		\draw[fill=blue,color=blue] (3.816684,2.007298) circle (\cirrad);
		\draw[fill=blue,color=blue] (4.888244,3.208098) circle (\cirrad);

		\draw[ thick,->] (-5,0) -- (-2,0);
		\draw[ thick,->] (-5,0) -- (-5,3);
		\node at (-5,3.5) {$\boldsymbol{z_c}$};

		
		\node at (-1.5,0) {$\boldsymbol{x_c}$};
	
		\end{scope}
		
	
		\end{scope}
		\end{tikzpicture}%
	\caption{An example of point sets used in the experiments, as well as the latent, true camera pose.}
	\label{fig:expr_example}
\end{figure}

We have compared the result of SHAPE to the results of minimising the reprojection error measured with the $\ell_2$ and $\ell_\infty$ norms. For the $\ell_2$-norm, the cost function is non-convex, resulting in multiple minima. However, we have calculated the global minimum using a brute-force strategy, in order to justify the selected criteria and not the specific methods.

\begin{figure}[t!]
		\centering{
			
			\begin{tikzpicture}[scale=0.8]
			\begin{axis}[
			xlabel=\large{Number of points},
			ylabel={\large{Squared error ($m^2$)}},
			xmin=5,ymin=0,
			legend style={fill=none},yticklabel style={
				/pgf/number format/fixed,
				/pgf/number format/precision=5
			},
			scaled y ticks=true
			]

			\addplot[thick,color=green,mark=+] coordinates {
	(5.000000,0.000991) (6.000000,0.000115) (7.000000,0.000055) (8.000000,0.000024) (9.000000,0.000014) (10.000000,0.000011) (11.000000,0.000008) (12.000000,0.000006) (13.000000,0.000004) (14.000000,0.000002) (15.000000,0.000001)
	  	 			
			};
			\addlegendentry{SHAPE}
			

			\addplot[thick,color=red,mark=.] coordinates {
				
		(5.000000,0.000945) (6.000000,0.000599) (7.000000,0.000459) (8.000000,0.000450) (9.000000,0.000266) (10.000000,0.000185) (11.000000,0.000187) (12.000000,0.000165) (13.000000,0.000172) (14.000000,0.000105) (15.000000,0.000092)
				 
			};
			\addlegendentry{$\ell_2$-norm minimisation}
			

			\addplot[thick,color=blue,mark=o] coordinates {
	(5.000000,0.001551) (6.000000,0.001057) (7.000000,0.000997) (8.000000,0.001123) (9.000000,0.000381) (10.000000,0.000202) (11.000000,0.000198) (12.000000,0.000198) (13.000000,0.000198) (14.000000,0.000138) (15.000000,0.000141)
			};
			\addlegendentry{$\ell_\infty$-norm minimisation}
			
			
			\end{axis}
			\end{tikzpicture}
		}
	\caption{Results of camera location estimation with fixed orientation.}
	\label{fig_or_known}
\end{figure}
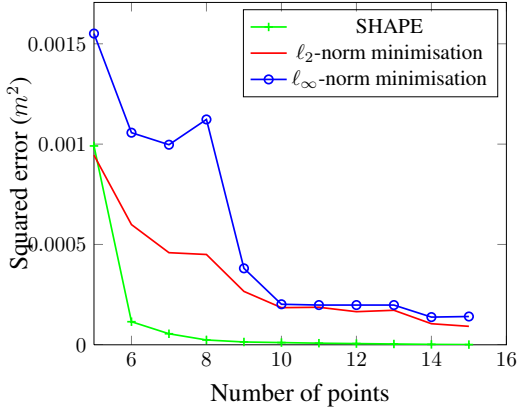


Figure \ref{fig_or_known} depicts the averaged result of camera pose estimation using the three approaches.  Here we have assumed that the camera orientation is known and given as an input to the algorithms. Although initially, i.e. for a small number of correspondences, the results of SHARP are worse than the norm-based methods, our algorithm converges much faster and the difference in accuracy becomes more evident as the number of correspondences increases. Moreover, we can see that the error of SHARP converges to $0$, which is not the case for norm-based approaches. 

\begin{figure}[t!]
\centering{
	
	\begin{tikzpicture}[scale=0.8]
	\begin{axis}[
	xlabel=\large{Number of points},
	ylabel={\large{Squared error ($m^2$)}},
	xmin=5,ymin=0,
	legend style={fill=none}
	]

	\addplot[thick,color=green,mark=+] coordinates {
	 (5.000000,0.004389) (6.000000,0.001607) (7.000000,0.000703) (8.000000,0.000317) (9.000000,0.000240) (10.000000,0.000150) (11.000000,0.000089) (12.000000,0.000047) (13.000000,0.000030) (14.000000,0.000021) (15.000000,0.000014)
	 	};
	\addlegendentry{SHAPE}
	
	
	\addplot[thick,color=red,mark=.] coordinates {
	 (5.000000,0.046367) (6.000000,0.013563) (7.000000,0.012641) (8.000000,0.010149) (9.000000,0.007432) (10.000000,0.006764) (11.000000,0.004571) (12.000000,0.002951) (13.000000,0.002820) (14.000000,0.002223) (15.000000,0.002531)
	 	};
	\addlegendentry{$\ell_2$-norm minimisation}
	

	\addplot[thick,color=blue,mark=o] coordinates {
		 (5.000000,0.072764) (6.000000,0.025588) (7.000000,0.021578) (8.000000,0.017318) (9.000000,0.015199) (10.000000,0.012907) (11.000000,0.005017) (12.000000,0.004468) (13.000000,0.006510) (14.000000,0.006510) (15.000000,0.004128)
	};
	\addlegendentry{$\ell_\infty$-norm minimisation}
	
	
	\end{axis}
	\end{tikzpicture}
}
	\caption{Results of full camera pose estimation.}
	\label{fig_or_unknown}
\end{figure}
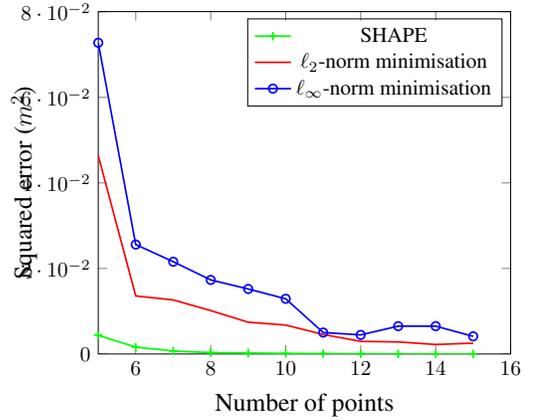

Figure \ref{fig_or_unknown} depicts the pose estimation results for the case of unknown camera orientation. We can see that similar results apply to this case as well.

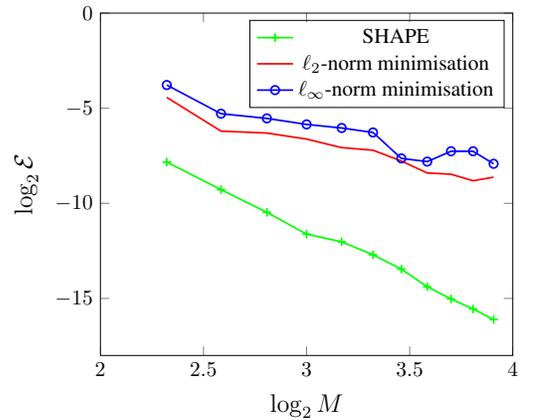
\begin{figure}[t!]
	\centering{
		
		\begin{tikzpicture}[scale=0.8]
		\begin{axis}[
		xlabel=\large{$\log_2 M$},
		ylabel={\large{$\log_2 \mathcal{E}$}},
		xmin=2,ymin=-18,ymax=0,
		xmax=4,
		legend style={fill=none}
		]

	\addplot[thick,color=green,mark=+] coordinates {
	 (2.321928,-7.832006) (2.584963,-9.281330) (2.807355,-10.474428) (3.000000,-11.621650) (3.169925,-12.023118) (3.321928,-12.705398) (3.459432,-13.461768) (3.584963,-14.385775) (3.700440,-15.029595) (3.807355,-15.554497) (3.906891,-16.108816)
	 
	};
	\addlegendentry{SHAPE}
	
	
	\addplot[thick,color=red,mark=.] coordinates {
		 (2.321928,-4.430745) (2.584963,-6.204221) (2.807355,-6.305740) (3.000000,-6.622581) (3.169925,-7.071956) (3.321928,-7.207924) (3.459432,-7.773401) (3.584963,-8.404568) (3.700440,-8.469989) (3.807355,-8.813370) (3.906891,-8.626075)
	};
	\addlegendentry{$\ell_2$-norm minimisation}
	

	\addplot[thick,color=blue,mark=o] coordinates {
	 (2.321928,-3.780639) (2.584963,-5.288407) (2.807355,-5.534307) (3.000000,-5.851612) (3.169925,-6.039839) (3.321928,-6.275658) (3.459432,-7.638903) (3.584963,-7.806234) (3.700440,-7.263178) (3.807355,-7.263178) (3.906891,-7.920381)
	 
	};
	\addlegendentry{$\ell_\infty$-norm minimisation}
	

		\end{axis}
		\end{tikzpicture}
	}
	\caption{Log-log plot visualising the convergence rate of different algorithms.}
	\label{fig_or_loglog}
\end{figure}

To have a better visualisation of the convergence rate of the algorithms,  we have additionally depicted a log-log plot of the error values, in Figure \ref{fig_or_loglog}. In the log-log plot, convergence rates correspond to the gradients. Therefore, it is easily verified that the error decay rate of the SHAPE algorithm is faster than other approaches. 

\section{Conclusion}

We have proposed the SHAPE algorithm, a fundamentally novel approach toward solving the PnP or camera pose estimation problem, using consistency regions and half-space intersections. We showed that SHAPE converges to zero error as the number of point correspondences tends to infinity.  Moreover, we have shown that our algorithm benefits from a linear time complexity.
 
Further work needs to be done to handle incorrect point correspondances and uncertainty in the feature point locations. We can develop novel outlier-detection techniques to remove points with large error and then increase the pixel width for smaller error values. Another possible extensions would be to incorporate other camera models and also relax some known parameters of the problem, such as the focal length or projections of some points. 

\bibliographystyle{IEEEbib}
\bibliography{refs}
\end{document}